# Power Market Price Forecasting via Deep Learning

Yongli Zhu, Renchang Dai, Guangyi Liu, Zhiwei Wang
GEIRI North America
Graph Computing and Grid Modernization Department
San Jose, California
yzhu16@vols.utk.edu

Songtao Lu
University of Minnesota Twin Cities
Department of Electrical and Computer Engineering
Minneapolis, Minnesota
lus@umn.edu

*Abstract*—A study on power market price forecasting by deep learning is presented. As one of the most successful deep learning frameworks, the LSTM (Long short-term memory) neural network is utilized. The hourly prices data from the New England and PJM day-ahead markets are used in this study. First, a LSTM network is formulated and trained. Then the raw input and output data are preprocessed by unit scaling, and the trained network is tested on the real price data under different input lengths, forecasting horizons and data sizes. Its performance is also compared with other existing methods. The forecasted results demonstrate that, the LSTM deep neural network can outperform the others under different application settings in this problem.

## I. INTRODUCTION

Nowadays, power markets have been playing more and more important roles in the deregulated power systems. As a strategic information, the power market price, aka. market clearing price, possesses vital significance in market design, trading strategy developing and system planning.

An accurate power market price forecasting is crucial to optimal operation and planning for both regional power utilities and ISO (Independent System Operator) companies. The forecasted price will be useful in making decisions, e.g. which generators to bid. It in turn also affects the final market clearing prices. Forecasting techniques are also required for modeling the market dynamics and pricing the power energy derivatives like futures or options, which are common practices in the trading risk control.

Among the existed machine learning methods, the deep leaning method can be regarded as one of the most successful tools so far, especially in image and text mining areas. However, its application for time series in power systems is less reported. In [1] the multi-channel deep convolutional neural networks were devised. This architecture extracts features by using CNN (Convolutional Neural Networks) on each input (1-D time-series) separately and feeds them to a conventional Multilayer Perceptron Neural Network (MLP NN) for classification. In this way CNN is used on temporal-coherent signals instead of using the whole input time series vector for single-channel. Similar usage pattern for classification is described in [2], where a so-called multi-scale technique is applied to consider the fact that a time series often has features at different time scales, thus adding a multi-branch layer to extracts features at different scales and /or frequencies might obtain a better feature representation.

While the classification problem has been treated by many researchers [3], there is still little literature on the time series forecasting application by deep learning frameworks. In [4], a short-term wind forecasting network is implemented by using CNN as the front end to learn the latent features and feeding them into the next level MLP NN for predicting. For the multivariate time series regression, using CNN to estimate the remaining useful life of system components is reported [5]. Inspired by the success of deep learning architectures in computer vision, some researcher [6] innovatively encodes the time series in 2-D images as input for CNN to identify the deep structure in time series. However, the overhead costs of time and storage due to encoding/decoding limit its application in real time for large datasets.

Compared to the aforementioned CNN, another kind of widely used deep leaning framework, i.e. RNN (Recurrent Neural Network) is more suitable for time series applications. The motivation of this paper is to apply one of the most successful RNN structure, i.e. LSTM (Long Short-Term Memory), into the power market price forecasting problem. This paper is organized as follows. In Section II, the original price data is preprocessed by unit scaling for the convenience of training; the LSTM-based deep learning architectures are illustrated in Section III. Section IV outlines of the LSTM-based price forecasting framework with emphasis on the network structure design and hyperparameter settings. Forecasted results are then compared with other machine learnings methods, i.e. SVM (Support Vector Machine) [7] and DT (Decision Tree). Section V summarizes the contributions and future research directions.

## II. DATA WRANGLING

### A. Data description

The power market price data used in this paper are from the New England ISO (ISO-NE) and PJM in the year-2009 [8], [9]. The data is hourly based which is used for day-ahead power market.

The original data scale can be as large as 500 MWh/$. This is not favorable for forecasting due to some potential numerical issue. Thus, the next step is preprocessing, i.e. standardization.

### B. Preprocessing

The standardization of datasets is a common requirement for many machine-learning procedures. A usual way is normalization, i.e. to transform the data to its center by removing the mean value of each feature, then scale it by dividing their standard deviation (std) for each feature:

$$x_{std} = \frac{x - x_{mean}}{\sigma_x} \quad (1)$$

where, $x_{mean}$ represents the mean of the corresponding feature data column $x$, and $\sigma_x$ denotes the standard deviation.

An alternative way is to scale the features in between a given minimum and maximum value, e.g. in [0,1], so that the maximum absolute value of each feature is scaled to one:

$$x_{scaled} = \frac{x - x_{min}}{x_{max} - x_{min}} \quad (2)$$

where, $x_{min}$ and $x_{max}$ stand for the minimum and maximum value for each feature (column data) respectively.

Note that, during testing phrase in the latter part of this paper, the standardization is applied on both training and testing data; however, the information used for (e.g. the min and max values in (2)) are only calculated based on the training dataset to guarantee a genuine forecasting, because in reality the testing data can never be known in advance.

In this paper, the following way is adopted. Motivations to use this scaling for the power market price are:

1) Robust to very small standard deviations of features.

2) The price time series in power markets trading is not necessarily stationary; thus, scaling the test data using (1) based on the mean and std values from the training data set may incur large errors when transforming them back to the original scale; on the other hand, the power market trading is regulated by ISO based on certain grid-operation-codes, which typically confine the trading prices to some reasonable ranges, i.e. the min and max values can be assumed stationary in a local power market.

## III. PRINCIPLES OF DEEP LEARNING FOR FORECASTING

### A. Deep Learning Introduction

Nowadays, deep learning has achieved a huge success in various areas such as natural language processing, speech identification and synthesis, image recognition, auto-driving, etc. The two main deep learning architectures are CNN and RNN. Both have included a big family of variants. Here, the most representative variant for RNN will be introduced in the following paragraphs, i.e. the LSTM neural network. CNN is more suitable to the problem with some spatial correlation like image data [10]; while RNN (LSTM) is more flexible to handle time series with temporal dynamics due to its sequential inputting nature [11]. Thus, in this study, LSTM is chosen as the power market price forecaster.

### B. LSTM introduction

LSTM was introduced by Sepp Hochreiter and Jürgen Schmidhuber in 1997 [12]. Unlike traditional RNNs, an LSTM network is well-suited to learn from experiences to identify and predict the time series when there are very long-time lags with unknown size. This is one of the main reasons why LSTM outperforms alternative RNNs and other sequence learning methods like HMM (Hidden Markov Models) in certain types of problems. A basic RNN is shown in Fig. 1:

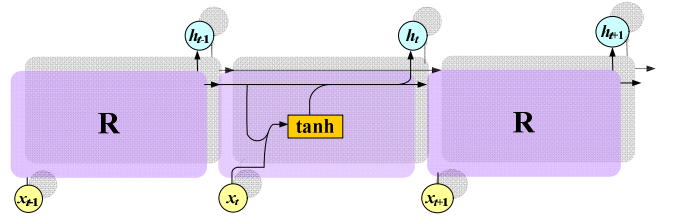

Figure 1. Basic structure of RNN

In the above diagram, each rectangular box "R" stands for a RNN unit. The input at step-$t$ is $x_t$ and the hidden layer function (*tanh* function) outputs a value $h_t$ passing from one step of the network to the next.

One of the drawbacks of traditional RNNs is the problem of "Long-Term Dependencies", i.e. with the time steps growing, the calculated gradient will either be exploding or vanishing drastically during training process.

Hence, LSTM is designed to avoid the above long-term dependency problem. Unlike standard RNNs, its repeating module has a more complicated inner structure: there exist more than one hidden layer function as shown in Fig. 2

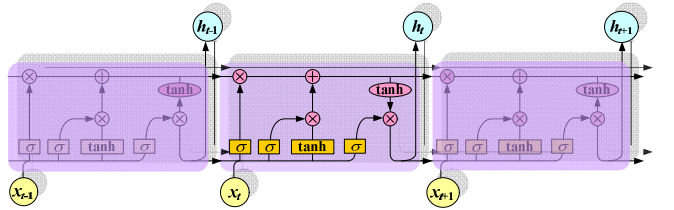

Figure 2. Basic structure of LSTM

The innovations of LSTM are the two concepts: "gate" ($\sigma$) and cell state ($s_t$). Gates are used to optionally allow information to pass through. Mathematically, it is an activation function (e.g. *sigmoid*) with pointwise multiplication operations. The gates output values between 0 and 1, describing the portion of information to be remembered. A value of 0 means "forget", while a value of 1 means "remember". An LSTM has three kinds of these gates: a forget gate $f_t$, an input gate $i_t$ and an output gate $o_t$, to control the cell states of step-$t$. Fig.3 shows the internal structure of one LSTM cell unit.

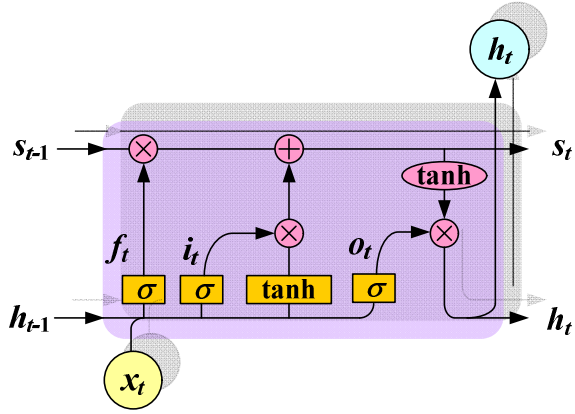

Figure 3. The internal structure of one LSTM cell unit

The updating equations used in the above diagram are:

$$f_t = \sigma(W_f x_t + U_f h_{t-1} + b_f) \quad (3)$$
$$i_t = \sigma(W_i x_t + U_i h_{t-1} + b_i) \quad (4)$$
$$o_t = \sigma(W_o x_t + U_o h_{t-1} + b_o) \quad (5)$$
$$s_t = f_t \circ s_{t-1} + i_t \circ \sigma(W_s x_t + U_s h_{t-1} + b_s) \quad (6)$$
$$h_t = o_t \circ \tanh(s_t) \quad (7)$$

where, "$\circ$" stands for the inner product operation and:

$x_t$: input of step $t$

$s_t$: cell state of step $t$

$h_t$: output of step $t$

$\sigma$: activation function (*sigmoid*)

$W_k, U_k, b_k$: weight matrices or vectors ($k = f, i, o, s$)

$f_t, i_t, o_t$: different gate outputs of step $t$

## IV. RESULTS AND DISCUSSION

In this section, three experiments are carried out: 1) the first and second experiments respectively investigate the forecasting performance of the deep learning method under different input lengths of features and different forecasting horizons on the same New England ISO (ISO-NE) dataset. 2) the last experiment is to test the forecasting performance on a small size dataset of PJM market. In the meantime, the comparison with other forecasting methods are also presented. All the experiments here are implemented in TensorFlow [13].

### A. Study for Different Input Length

2/3 of this year-2009 data is used for training and the remaining 1/3 is served for testing. The forecasting workflow is as follows. The data is depicted in Fig. 4.

■ Feature Engineering

- Network inputs: the inputs are the past prices in a certain time window of length $L$ (look back window), i.e. $x(t-1), x(t-2), ..., x(t-L)$; in this study, length $L = 4, 8, 12, 24$ are compared;

- Network outputs: the forecasted price value at time $t$, i.e. $x(t)$.

■ Network design

- Hidden Layer number: 1; LSTM cell units: 4

- Batch size: 1 (the number of training examples in one forward/backward process. The higher the batch size, the more memory space is needed).

- Learning rate = 0.001.

- Loss Function: The $L_2$ loss is considered.

■ Training and Testing

- Training epochs: 50 (more epochs might bring potential accuracy improvement but longer time)

- Cross validation: 2/3 for training; 1/3 for testing.

Hence, the designed LSTM network has an input layer with $L$ inputs, a hidden layer with four LSTM blocks and an output layer of dimension one. The default sigmoid activation function is used for each LSTM block. The network is trained over 50 epochs where a batch size of one is used.

The RMSE (root mean square error) results are listed in Table 1 to Table 3 for each method. The comparison plot for different input lengths is shown in Fig. 5.

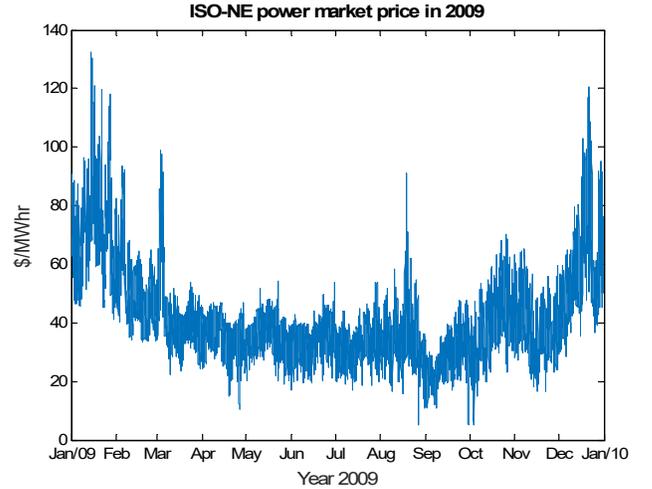

Figure 4. The year-2009 power market price data of ISO-NE

TABLE I. RESULTS BY DECISION TREE

| Performance ($/MWhr) | Different Input Lengths | | | |
|---|---|---|---|---|
| | DT L=4 | DT L=8 | DT L=12 | DT L=24 |
| *Test RMSE* | 5.78 | 5.71 | 5.47 | 5.40 |
| *Train RMSE* | 0.02 | 0.01 | 0.01 | 0.01 |

TABLE II. RESULTS BY SVM

| Performance ($/MWhr) | Different Input Lengths | | | |
|---|---|---|---|---|
| | SVM L=4 | SVM L=8 | SVM L=12 | SVM L=24 |
| *Test RMSE* | 5.99 | 5.46 | 5.50 | 5.16 |
| *Train RMSE* | 5.45 | 4.92 | 4.96 | 4.73 |

TABLE III. RESULTS BY LSTM

| Performance ($/MWhr) | Different Input Lengths | | | |
|---|---|---|---|---|
| | LSTM L=4 | LSTM L=8 | LSTM L=12 | LSTM L=24 |
| Test RMSE | 4.48 | 4.45 | 4.35 | 3.83 |
| Train RMSE | 3.71 | 3.62 | 3.58 | 3.02 |

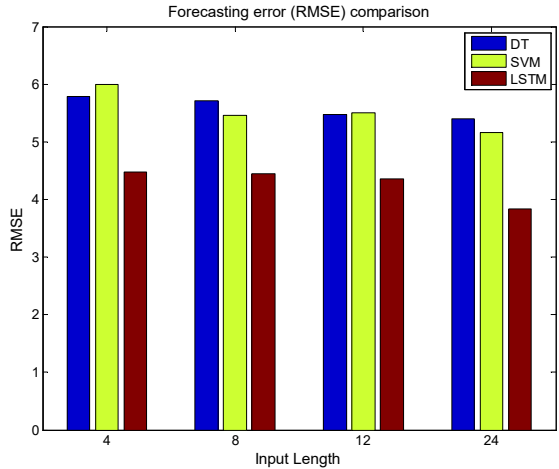

Figure 5. RMSE comparisons for different input lengths

Forecasted prices are illustrated in Figure. 6.

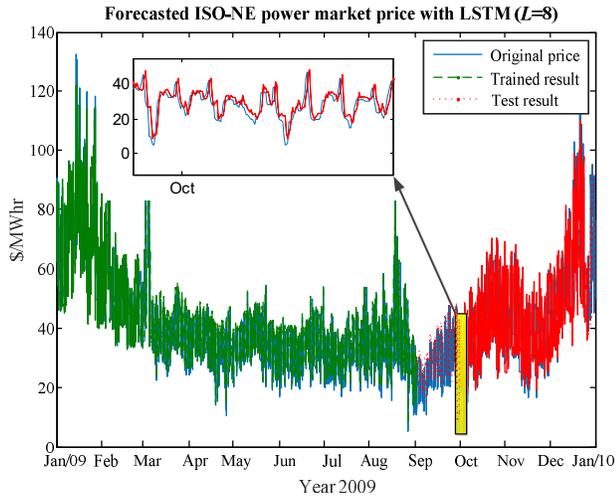

Figure 6. Training and testing results by LSTM

### B. Study for Different Forecasting Horizons

The previous forecasting experiment is for the one-hour ahead prediction. To strictly inspect the forecasting performance of the deep learning method, results under other forecasting horizons ("$H$") are summarized in Table 4 to Table 6. The input length $L$ for each case here is fixed to 24. The RMSE comparison is shown in Fig. 7.

TABLE IV. RESULTS BY DECISION TREE

| Performance ($/MWhr) | Different Forecasting Horizons | | | |
|---|---|---|---|---|
| | DT H=3 | DT H=6 | DT H=12 | DT H=24 |
| Test RMSE | 7.93 | 8.861 | 9.861 | 10.813 |
| Train RMSE | 0.01 | 0.01 | 0.009 | 0..09 |

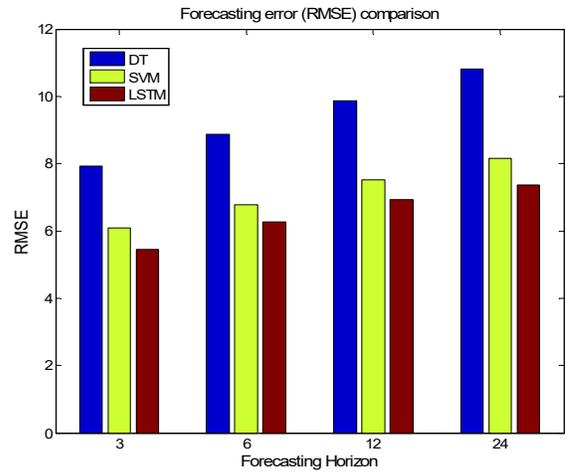

Figure 7. RMSE comparisons for different forecasting horizons

TABLE V. RESULTS BY SVM

| Performance ($/MWhr) | Different Forecasting Horizons | | | |
|---|---|---|---|---|
| | SVM H=3 | SVM H=6 | SVM H=12 | SVM H=24 |
| Test RMSE | 6.10 | 6.78 | 7.52 | 8.15 |
| Train RMSE | 5.38 | 6.01 | 6.61 | 7.53 |

TABLE VI. RESULTS BY LSTM

| Performance ($/MWhr) | Different Forecasting Horizons | | | |
|---|---|---|---|---|
| | LSTM H=3 | LSTM H=6 | LSTM H=12 | LSTM H=24 |
| Test RMSE | 5.45 | 6.27 | 6.93 | 7.37 |
| Train RMSE | 4.76 | 5.39 | 5.86 | 6.48 |

### C. Study for Insufficient Data

The PJM hourly price in 2009 June is shown in Fig. 8.

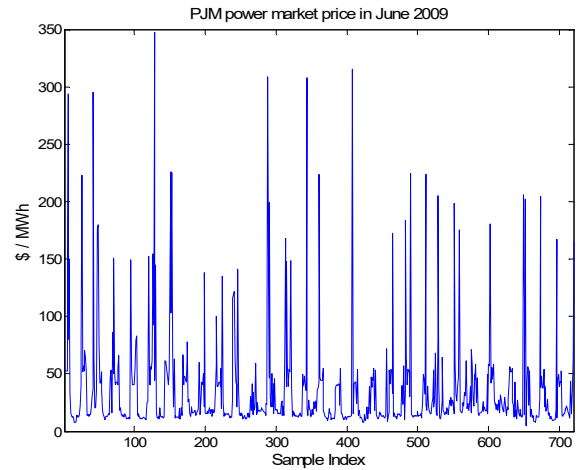

Figure 8. The Power market price data of PJM in June. 2009

This dataset has only 720 samples. Likewise, 2/3 of it are used for training and 1/3 for testing. Similar to previous sections, here the results under different forecasting horizons are shown in Table. 7 to Table 9. The input length $L$ for each case is fixed to 24. The RMSE comparison plot is in Fig. 9.

TABLE VII. RESULTS BY DECISION TREE

| Performance ($/MWhr) | Different Forecasting Horizons | | | |
|---|---|---|---|---|
| | DT H=3 | DT H=6 | DT H=12 | DT H=24 |
| Test RMSE | 57.203 | 54.346 | 60.947 | 49.153 |
| Train RMSE | 0.031 | 0.030 | 0.033 | 0.029 |

TABLE VIII. RESULTS BY SVM

| Performance ($/MWhr) | Different Forecasting Horizons | | | |
|---|---|---|---|---|
| | SVM H=3 | SVM H=6 | SVM H=12 | SVM H=24 |
| Test RMSE | 37.08 | 35.67 | 36.56 | 36.05 |
| Train RMSE | 42.59 | 43.03 | 43.57 | 42.13 |

TABLE IX. RESULTS BY LSTM

| Performance ($/MWhr) | Different Forecasting Horizons | | | |
|---|---|---|---|---|
| | LSTM H=3 | LSTM H=6 | LSTM H=12 | LSTM H=24 |
| Test RMSE | 37.38 | 35.55 | 36.05 | 37.51 |
| Train RMSE | 38.94 | 39.93 | 39.56 | 36.76 |

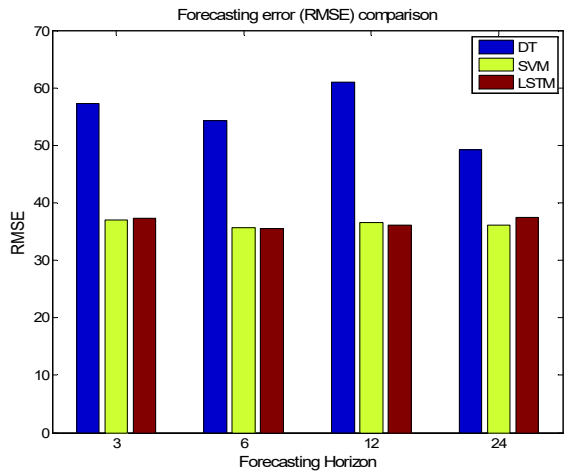

Figure 9. RMSE comparisons for different forecasting horizons (PJM data)

*D. Discussion*

In subsection A, both DT and SVM are outperformed by the deep learning method in terms of the RMSE. Another observation is that the performance improvements brought by increasing the input length is not significant. Thus, a reasonable long input sequence can be acceptable.

In subsection B, the performance deteriorates for each method when *H* increases. Although the deep leanring method still win over DT and SVM, DT shows its drawback in easily overfittings, i.e., it has extremely low training errors (e.g. 0.01) but much larger testing errors. The LSTM can avoid such overfitting by adopting some techniques in the training stage, such as the "*dropout*" method [9].

In subsection C, the advantage of SVM in good generalization ability on small size data can be observed. As a well-known fact, deep learning methods usually perform better on big enough datasets. Even though, here its performance can still ravel that of SVM.

To sum up, the LSTM-based deep learning framework can successfully forecast the power market price with satisfactory accuracy under different input lengths, forecasting horizons and data sizes.

V. CONCLUSION

In this paper, a LSTM deep neural network is devised for power market price forecasting, which achieves a good performance with a black-box style training process. The next step is to elaborate an effective way to tune the network parameters (layer numbers, cell unit numbers, etc.) for higher accuracies based on theoretical analysis. Research regarding combining deep learning with signal decomposition methods [14] to enhance the forecasting capability is also undergoing.

ACKNOWLEDGMENT

This work is supported by the State Grid Corporation technology project SGRIJSKJ (2016) 800.